\documentclass{article}
\usepackage{ijcai22}

\usepackage{times}
\usepackage{soul}
\usepackage{url}
\usepackage[utf8]{inputenc}
\usepackage[small]{caption}
\usepackage{graphicx}
\usepackage{amsmath}
\usepackage{amsthm}
\usepackage{booktabs}
\usepackage{algorithm}
\usepackage{algorithmic}

\usepackage{hyperref}
\usepackage{graphicx}
\usepackage{wrapfig}
\usepackage{multirow}
\usepackage{wrapfig}
\usepackage{bm}
\usepackage[table,xcdraw]{xcolor}
\usepackage{threeparttable}
\usepackage{siunitx}

\usepackage{amsmath,amsfonts}
\usepackage{array}

\usepackage{textcomp}
\usepackage{stfloats}
\usepackage{verbatim}
\usepackage{amssymb}

\usepackage{bibspacing}
\title{Graph Pooling for Graph Neural Networks: Progress, Challenges, and Opportunities}

\author{
Chuang Liu$^1$\thanks{This work was done when Chuang Liu worked as an intern at JD Explore Academy.}
\and
Yibing Zhan$^2$\and
Jia Wu$^3$\and
Chang Li$^2$\and
Bo Du$^4$\and
Wenbin Hu$^1$\footnote{Corresponding Author}\and \\
Tongliang Liu$^5$\And
Dacheng Tao$^5$
\affiliations
$^1$School of Computer Science, Wuhan University, Wuhan, China\\
$^2$JD Explore Academy, JD.com, China \\
$^3$School of Computing, Macquarie University, Sydney, Australia\\
$^4$National Engineering Research Center for Multimedia Software, Wuhan University, Wuhan, China \\
$^5$School of Computer Science,  University of Sydney, Sydney, Australia
\emails
\{chuangliu, dubo, hwb\}@whu.edu.cn,
\{zhanyibing, lichang93\}@jd.com,
jia.wu@mq.edu.au
tongliang.liu@sydney.edu.au,
dacheng.tao@gmail.com
}
\begin{document}

\maketitle

% A Survey on Complex Knowledge Base Question Answering: Methods, Challenges and Solutions. 4483-4491
% Cross-Domain Recommendation: Challenges, Progress, and Prospects. 4721-4728
% Deep Learning for Community Detection: Progress, Challenges and Opportunities. 4981-4987

\begin{abstract}

Graph neural networks have emerged as a leading architecture for many graph-level tasks, such as graph classification and graph generation. As an essential component of the architecture,  graph pooling is indispensable for obtaining a holistic graph-level representation of the whole graph. Although a great variety of methods have been proposed in this promising and fast-developing research field, to the best of our knowledge, little effort has been made to systematically summarize these works. To set the stage for the development of future works, in this paper, we attempt to fill this gap by providing a broad review of recent methods for graph pooling. Specifically, 1) we first propose a taxonomy of existing graph pooling methods with a mathematical summary for each category; 2) then, we provide an overview of the libraries related to graph pooling, including the commonly used datasets, model architectures for downstream tasks, and open-source implementations; 3) next, we further outline the applications that incorporate the idea of graph pooling in a variety of domains; 4) finally, we discuss certain critical challenges facing current studies and share our insights on future potential directions for research on the improvement of graph pooling.
  
 %  we present an additional mathematical summary of the existing works in a unified framework. 
 
 % provide an in-depth understanding of graph pooling, we
  
\end{abstract}

\section{Introduction}

%\begin{figure}[!t] % !htb
%\begin{center}
%\setlength{\abovecaptionskip}{-0.5cm}   %调整图片标题与图距离
%\setlength{\belowcaptionskip}{-0.9cm}   %调整图片标题与下文距离
%\includegraphics[width=0.99\linewidth]{pics/pool.eps}
%\end{center}
%\caption{An illustrative example of graph pooling.}
%\label{fig:survey}
%\end{figure}
Graph Neural Networks (GNNs) have achieved a substantial improvement over many graph-level tasks such as graph classification~\cite{fair-graph-classification}, graph regression~\cite{mincut}, and graph generation~\cite{gmt}. Specifically, GNNs have been successfully applied to graph-level tasks across a broad range of areas such as chemistry and biology~\cite{gmt}, social network~\cite{Anomaly-Detection-survey}, computer vision~\cite{splinecnn}, natural language processing~\cite{gpool-text}, and recommendation~\cite{user-as-graph}. 

\begin{figure*}[!t] % !htb
\begin{center}
\setlength{\abovecaptionskip}{-0.3cm}   %调整图片标题与图距离
\setlength{\belowcaptionskip}{-0.5cm}   %调整图片标题与下文距离
\includegraphics[width=0.8\linewidth]{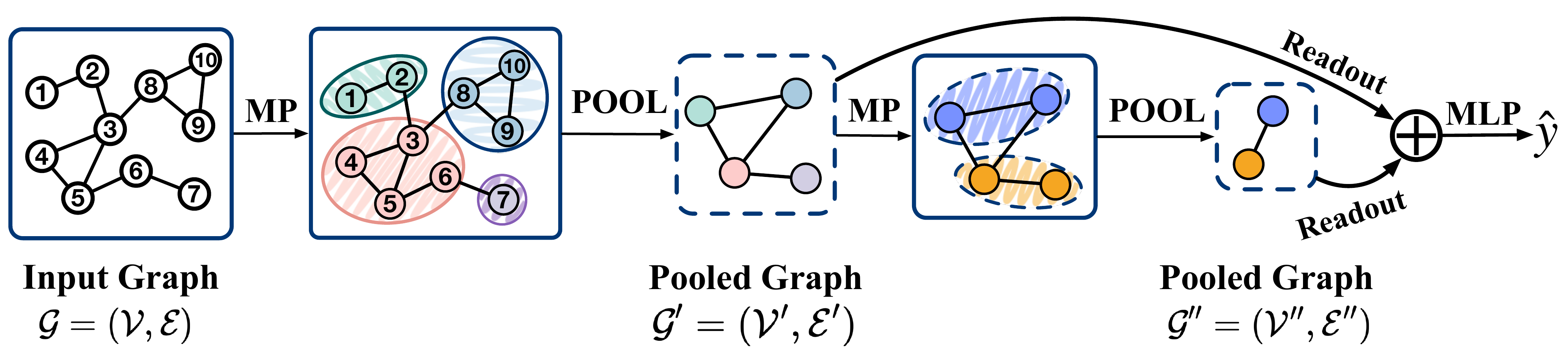}
\end{center}
\caption{An illustrative example of graph pooling. Here, MP refers to message passing, and POOL refers to the pooling function.}
\label{fig:survey}
\end{figure*}

Different from node-level tasks, which mainly use the graph convolutional network (GCN)~\cite{gcn} to generate node representations for downstream tasks,  graph-level tasks require holistic graph-level representations for graph-structured inputs whose size and topology are varying.  Therefore, for graph-level tasks, the pooling mechanism is an essential component, which condenses the input graph with node representations generated by GCN into a smaller sized graph or a single vector, as shown in Figure~\ref{fig:survey}. 

%Also, graph pooling enables us to deal with graph inputs with variable sizes and structures in the datasets.

In order to obtain an effective and reasonable graph representation, many designs of graph pooling have been proposed, which could be roughly divided into Flat Pooling (\textbf{Section~\ref{sec:fpooling}}) and Hierarchical Pooling (\textbf{Section~\ref{sec:hpooling}}). The former directly generates a graph-level representation in one step, mostly taking the average or sum over all node embeddings as the graph representation~\cite{duvenaud}, while the latter coarsens a graph gradually into a smaller sized graph by two main means:  Node Clustering Pooling and Node Drop Pooling. Specifically, node clustering pooling~\cite{diffpool} groups nodes into clusters as a coarsened graph, which is time-and space-consuming~\cite{mincut}. In contrast, node drop pooling~\cite{graph-u-net} selects a subset of nodes from the original graph to construct a coarsened graph, which is more efficient and more suitable for large-scale graphs~\cite{sagpool} but suffers from inevitable information loss~\cite{ipool}.   

% graph clustering~\cite{Community-Detection-survey-1,Community-Detection-survey-2},

Although such state-of-the-art graph pooling methods have been proposed, only a few recent works have attempted to comprehensively evaluate the effects of graph pooling~\cite{rethink-pooling,understanding-pooling}, and a systematic review of the progress of and challenges facing this emerging area is still lacking. To fill the gaps, we comprehensively survey graph pooling in this paper, including proposing a taxonomy and formulating relevant frameworks (\textbf{Section~\ref{sec:methods}}), overviewing libraries (\textbf{Section~\ref{sec:library}}), outlining applications (\textbf{Section~\ref{sec:application}}), and discussing future research directions (\textbf{Section~\ref{sec:challenge}}). To the best of our knowledge, our paper is the first attempt to present a systematic and comprehensive review of recent progress on graph pooling. The purpose of this paper is to provide new practitioners with a comprehensive understanding of graph pooling and to keep researchers informed about the latest advancements in this field.
% This paper is intended to help new practitioners gain a deep appreciation of graph pooling and keep researchers up to date with the developments of graph pooling. 

\section{Problem Formulation}

\paragraph{Notions.} Let $\mathcal{G}=(\mathcal{V}, \mathcal{E})$ denote a graph with node set $\mathcal{V}$ and edge set $\mathcal{E}$. Node features are denoted as $\boldsymbol{X} \in \mathbb{R}^{ n \times d}$, where $n$ is the number of nodes, and $d$ is the dimension of node features. The adjacency matrix is defined as  $\boldsymbol{A} \in \{0, 1\}^{ n \times n}$. $\boldsymbol{A}[i, j]=1$ if there exists an edge between node $v_{i}$ and node $v_{j}$, otherwise, $\boldsymbol{A}[i, j]=0$.

% \paragraph{Graph Neural Networks.} GNNs can learn node representations through aggregation scheme, which consists of two functions, $\operatorname{AGG}$regate and $\operatorname{COM}$bine :
% \begin{equation}
% \boldsymbol{h}_{v}^{(l+1)}=\operatorname{COM}^{(l)}\left(\left\{\boldsymbol{h}_{v}^{(l)}, \operatorname{AGG}^{(l)}\left(\left\{\boldsymbol{h}_{v^{\prime}}^{(l)}: v^{\prime} \in \mathcal{N}_{v}\right\}\right)\right\}\right),
% \end{equation}
% where $\boldsymbol{h}_{v}^{(l+1)} \in \mathbb{R}^{d}$ is the representation vector  for node $v$ in the $l$-layer, and $\mathcal{N}_{v}$ is the set of neighbors of node $v$. However, to obtain a representation of the entire graph for downstream graph-level tasks, an additional pooling operation is required.

\paragraph{Graph Pooling.}  Let a graph pooling operator be defined as any function $\operatorname{POOL}$ that maps a graph $\mathcal{G}$ to a new pooled graph $\mathcal{G}^{\prime}=(\mathcal{V}^{\prime}, \mathcal{E}^{\prime})$ :
\begin{equation}
\mathcal{G}^{\prime}=\operatorname{POOL}(\mathcal{G}),
\label{eq:pool}
\end{equation}
where $ |\mathcal{V}^{\prime}| < |\mathcal{V}| $~\footnote{In some very specific cases, there exits  $ |\mathcal{V}^{\prime}| \geq |\mathcal{V}| $, causing the graph to be upscaled by pooling.}. The primary goal of graph pooling is to reduce the number of nodes in a graph while preserving the semantic information of the graph.

\section{Approaches for Graph Pooling}
\label{sec:methods}

Graph pooling can be roughly divided into flat pooling and hierarchical pooling according to its role in graph-level representation learning. The former directly generates graph-level representations in a single step ($|\mathcal{V}^{\prime}| = 1$), while the latter coarsens the graph gradually into a smaller sized graph ($|\mathcal{V}^{\prime}| > 1$).

% mostly discarding the topological information,

\subsection{Flat Pooling}
\label{sec:fpooling}
Flat pooling, also known as graph readout operation, directly generates a graph-level representation $\boldsymbol{h}_{G}$ in one step. Thus, Eq.~\ref{eq:pool} in the case of flat pooling can be denoted as:
%(a single node instead of a new coarsened graph)
\begin{equation}
\boldsymbol{h}_{G}=\operatorname{POOL}_\text{flat}(\mathcal{G}),
\end{equation}
%\begin{equation}
%\boldsymbol{h}_{G}=\operatorname{POOL}\left(\left\{\boldsymbol{h}_{v} : v \in \mathcal{V}\right\}\right).
%\end{equation}
where $\operatorname{POOL}_\text{flat}$ denotes the graph pooling function, which must: \textbf{1)} output fixed-sized graph representations when input graphs are of different sizes; \textbf{2)} output the same representation when the order of nodes of an input graph changes. 

In light of the above discussions, several designs of flat pooling layers have been proposed. The most commonly used method is the sum-pool or mean-pool, which performs averaging or summing operations over all node representations~\cite{duvenaud,gin}. Some methods~\cite{deepsets,gfn,dropgnn} perform an additional non-linearity transformation to improve the expressive power of pooling methods. Moreover, some methods~\cite{gated,Diffusion-CNN,msnapool,self-attention,Multi-level} introduce the soft attention mechanism to determine the weight of each vertex in the final graph-level representation. Besides, some methods~\cite{sortpool,qsgcnn} apply convolutional neural networks to sorted node representations. Different from the above methods, which collect the first-order statistic information of node representations, SOPool, proposed by Wang \textit{et al.}~\shortcite{second-order}, considers the important second-order statistics, which refers to functions that utilize the second power of node features. Furthermore, DKEPool~\cite{dke-pool} takes into consideration the entire node distribution of a graph. Due to space limitation, some other flat pooling methods such as Set2set~\cite{set2set}, DEMO-Net~\cite{degree-pool}, SSRead~\cite{SSread}, and GMT~\cite{gmt} are not presented here. 

Most of the above flat pooling methods perform operations on node representations to obtain graph-level representations without consideration of the intrinsic hierarchical structures of graphs, which causes information loss and degrades the performance of graph representations~\cite{understand-attention,expressive-pooling}.

\subsection{Hierarchical Pooling}
\label{sec:hpooling}

Hierarchical pooling methods aim to preserve the hierarchical graph's structural information by iteratively coarsening the graph into a new graph in smaller size. The hierarchical pooling can be roughly classified into node clustering pooling, node drop pooling, and other pooling according to the manner in which it coarsens a graph.  The main difference between the first two types of methods is that node clustering pooling generates new nodes for the coarsened graph, whereas node drop pooling retains nodes from the original graph. Note that hierarchical pooling methods still technically employ flat pooling methods (readout in Figure~\ref{fig:survey}) to obtain the graph-level representation of the coarsened graph.

%A straightforward way to use hierarchical graph pooling for graph representation learning is to reduce the number of nodes to one. Then the resulted single vector is treated as the graph representation. 

%\begin{figure}[!t] % !htb
%\begin{center}
%\setlength{\abovecaptionskip}{-0.5cm}   %调整图片标题与图距离
%\setlength{\belowcaptionskip}{-0.9cm}   %调整图片标题与下文距离
%\includegraphics[width=0.99\linewidth]{pics/h-pool.png}
%\end{center}
%\caption{Overview of graph pooling. (a) Taxonomy of graph pooling operators. }
%\label{fig:hierarchical}
%\end{figure}

 \begin{table*}[!t]
\renewcommand\arraystretch{2.0} % 行间距
\setlength\tabcolsep{20pt} % 列间距
\centering
\begin{threeparttable}[b]
%\footnotesize
%\scriptsize
%\tiny
\resizebox{0.85\textwidth}{!}{%
\begin{tabular}{@{}p{3.6cm}<{\centering}|c|c|c@{}}
\toprule
\multicolumn{1}{c}{\Large \textbf{Models} }                                            & \multicolumn{1}{c}{\Large \textbf{CAM Generator} }                                                                                         & \multicolumn{1}{c}{\Large \textbf{Graph Coarsening} }                                                                                                                  & \multicolumn{1}{c}{\Large \textbf{Notes} }                                                                                                                                                                                                                                                                                                                                                                     \\ \midrule

%\begin{tabular}[c]{@{}l@{}}Graclus~\cite{graclus}\end{tabular}      & $\boldsymbol{C}=\left\{\mathbf{x}_{i}, \mathbf{x}_{j} \mid \arg \max _{j}\left(\frac{\mathbf{A}_{i j}}{\mathbf{D}_{i i}}+\frac{\mathbf{A}_{i j}}{\mathbf{D}_{j j}}\right)\right\}$  & $ \boldsymbol{X}^{\prime} =\boldsymbol{C}^{T} \boldsymbol{X} $                                                                                             & METIS                                                                                                                                                                                                                                                                                  \\ %\midrule

\begin{tabular}[c]{@{}l@{}}\Large  DiffPool\tnote{[1]}\end{tabular}     & $\boldsymbol{C}= \operatorname{softmax}( \operatorname{GNN}_\text{pool}( \boldsymbol{X},  \boldsymbol{A}) )$                                                    &  $\left\{\begin{aligned}&\boldsymbol{X}^{\prime} =\boldsymbol{C}^{T} \cdot \operatorname{GNN}_\text{emb}( \boldsymbol{X},  \boldsymbol{A}) \\ &\boldsymbol{A}^{\prime} = \boldsymbol{C}^{T} \boldsymbol{A} \boldsymbol{C} \\ \end{aligned}\right.$                                                                                              &  Auxiliary  Loss\tnote{ \textcircled{\raisebox{-0.9pt}{1}}} \\ \midrule

\begin{tabular}[c]{@{}l@{}}\Large NMF\tnote{[2]}\end{tabular}      & $\left\{\begin{aligned}& \boldsymbol{A} \approx \boldsymbol{U} \boldsymbol{V}  \\& \boldsymbol{C} = \boldsymbol{V}^{T} \end{aligned}\right.$ & $ \boldsymbol{X}^{\prime} =\boldsymbol{C}^{T} \boldsymbol{X}; \boldsymbol{A}^{\prime} = \boldsymbol{C}^{T} \boldsymbol{A} \boldsymbol{C} $                                                                                                 & --                                                                                                                                                                                                                                                                              \\ \midrule

\begin{tabular}[c]{@{}l@{}} \Large LaPool\tnote{[3]}\end{tabular}      & $\left\{\begin{aligned}& s_i=\|\sum_{j \in \mathcal{N}\left(v_i\right)} A_{i, j}\left(\mathbf{x}_i-\mathbf{x}_j\right)\|_2 \\ & \mathcal{V}_{c}=\left\{v_{i} \in \mathcal{V} \mid \forall v_{j}, s_{i}-\boldsymbol{A}_{i j} s_{j}>0\right\} \\ &\boldsymbol{C}=\operatorname{sparsemax}\left(\mathbf{p}\frac{\boldsymbol{X X}_{c}^{T}}{\|\boldsymbol{X}\|\left\|\boldsymbol{X}_{c}\right\|}\right)\end{aligned}\right.$ & $\left\{\begin{aligned}&\boldsymbol{X}^{\prime} = \operatorname{MLP}( \boldsymbol{C}^{T}   \boldsymbol{X})   \\ &\boldsymbol{A}^{\prime} = \boldsymbol{C}^{T} \boldsymbol{A} \boldsymbol{C} \\ \end{aligned}\right.$                                                                   & --                                                                                                                                                                                                                                                                                  \\ \midrule

\begin{tabular}[c]{@{}l@{}}\Large MinCut\tnote{[4]}\end{tabular}      & $\boldsymbol{C}=\operatorname{MLP}( \boldsymbol{X})$ & $ \left\{\begin{aligned}& \boldsymbol{X}^{\prime} =\boldsymbol{C}^{T} \boldsymbol{X}; \widehat{\boldsymbol{A}} = \boldsymbol{C}^{T}  \widetilde{\boldsymbol{A}} \boldsymbol{C}\\ & \boldsymbol{A}^{\prime} = \widehat{\boldsymbol{A}} - \boldsymbol{I}\text{diag}(\widehat{\boldsymbol{A}})  \end{aligned}\right. $                                                                                              & MinCut Loss\tnote{ \textcircled{\raisebox{-0.9pt}{2}}}                                                                                                                                                                                                                                                                                  \\ \midrule

\begin{tabular}[c]{@{}l@{}}\Large StructPool\tnote{[5]}\end{tabular}      & $\boldsymbol{C} : \operatorname{Minimiz} E(\boldsymbol{C}) $\tnote{ \textcircled{\raisebox{-0.9pt}{3}}} &    $ \boldsymbol{X}^{\prime} =\boldsymbol{C}^{T} \boldsymbol{X}; \boldsymbol{A}^{\prime} = \boldsymbol{C}^{T} \boldsymbol{A} \boldsymbol{C} $                                                                                             & --                                                                                                                                                                                                                                                                                  \\ \midrule

\begin{tabular}[c]{@{}l@{}}\Large MemPool\tnote{[6]}\end{tabular}      & $\left\{\begin{aligned}& \boldsymbol{C}_{i, j}=\frac{\left(1+\left\|\mathbf{x}_{i}-\mathbf{k}_{j}\right\|^{2} / \tau\right)^{-\frac{\tau+1}{2}}}{\sum_{j^{\prime}}\left(1+\|\mathbf{x}_{i}-\mathbf{k}_{j^{\prime}}\|^{2} / \tau\right)^{-\frac{\tau+1}{2}}} \\ & \boldsymbol{C}=\operatorname{softmax}\left(\Gamma \left(\overset{|m|}{  \underset{t=0} {\|}} \boldsymbol{C}_{t}\right)\right) \end{aligned}\right.$ & $\boldsymbol{X}^{\prime} = \operatorname{MLP}( \boldsymbol{C}^{T}   \boldsymbol{X})$                                                                   &  Auxiliary  Loss\tnote{ \textcircled{\raisebox{-0.9pt}{4}}}                                                                                                                                                                                                                                                                                \\ \midrule

\begin{tabular}[c]{@{}l@{}}\Large HAP\tnote{[7]}\end{tabular}      & $\left\{\begin{aligned}& \boldsymbol{T}= \operatorname{GCont} (\boldsymbol{X}) \\ & \boldsymbol{C}_{i j}=\sigma\left(\mathbf{p}^{T}[\boldsymbol{T}_{\text {Row}_{i}} \| \boldsymbol{T}_{\text{Col}_{j}}]\right) \\& \boldsymbol{C} = \operatorname{softmax}(\boldsymbol{C}) \end{aligned}\right.$ & $\left\{\begin{aligned}&\boldsymbol{X}^{\prime} = \operatorname{MLP}( \boldsymbol{C}^{T}   \boldsymbol{X}), \widehat{\boldsymbol{A}} = \boldsymbol{C}^{T} \boldsymbol{A} \boldsymbol{C}    \\ &\boldsymbol{A}^{\prime} = \operatorname{Gumbel-SoftMax}(\widehat{\boldsymbol{A}}) \\ \end{aligned}\right.$                                                                   & --                                                                                                                                                                                                                                                                                  \\ \midrule

\begin{tabular}[c]{@{}l@{}}\Large SEP\tnote{[8]}\end{tabular}      & $\boldsymbol{C} : \operatorname{Minimiz} \mathcal{H}^T(\mathcal{G}) $\tnote{ \textcircled{\raisebox{-0.9pt}{5}}} &    $ \boldsymbol{X}^{\prime} =\boldsymbol{C}^{T} \boldsymbol{X}; \boldsymbol{A}^{\prime} = \boldsymbol{C}^{T} \boldsymbol{A} \boldsymbol{C} $                                                                                             & --                                                                                                                                                                                                                                                                                  \\ %\midrule

\bottomrule
\end{tabular}%
}
%\begin{tablenotes}
%  \scriptsize  
%  \item[1] Auxiliary link prediction objective $L_{\mathrm{LP}}=\|A, C C^{T} \|_{F}$ and entropy regularization $L_{\mathrm{E}}=\frac{1}{n} \sum_{i=1}^{n} H\left(C_{i}\right)$, where H denotes the entropy function.
%  \end{tablenotes}
  \end{threeparttable}
  \begin{minipage}{0.85\linewidth} \scriptsize %\footnotesize %
 $^{[1]}$\cite{diffpool}; $^{[2]}$\cite{NMFPool}; $^{[3]}$\cite{lap-pool}; $^{[4]}$\cite{mincut}; $^{[5]}$\cite{structpool}; $^{[6]}$\cite{mem-pool}; $^{[7]}$\cite{hap}; $^{[7]}$\cite{sep}

 \textcircled{\raisebox{-0.9pt}{1}} Auxiliary loss consists of the link prediction objective loss and entropy regularization loss.   \textcircled{\raisebox{-0.9pt}{2}} MinCut loss consists of cut loss, which approximates the \textit{mincut} problem, and orthogonality loss, which spurs the assignments to be orthogonal.   \textcircled{\raisebox{-0.9pt}{3}} $E(\boldsymbol{C})$ is the Gibbs energy, which consists of unary energy and pairwise energy.  \textcircled{\raisebox{-0.9pt}{4}} Auxiliary loss is an unsupervised clustering loss, which spurs the model to learn clustering-friendly embeddings.  \textcircled{\raisebox{-0.9pt}{5}} $\mathcal{H}^T(\mathcal{G})$ is the  structural entropy for $\mathcal{G}$ on coding tree $T$. 

 \textbf{Notations:} $\boldsymbol{X}^{\prime} \in \mathbb{R}^{ c \times d}$ and  $\boldsymbol{A}^{\prime} \in \mathbb{R}^{ c \times c}$  are the adjacency matrix and feature matrix for the new graph, respectively; $\widetilde{ \boldsymbol{A}} \in \mathbb{R}^{ n \times n}$  is the adjacency matrix with self loop;  $\boldsymbol{L} \in \mathbb{R}^{ n \times n} $ is the graph Laplacian; $\boldsymbol{I}_{k}\in \mathbb{R}^{c \times c}$ is the identity matrix; $\mathbf{k} \in \mathbb{R}^{d}$ is a memory key vector; $\mathbf{p} \in \mathbb{R}^{ 1 \times 2d}$ is a trainable vector; $\tau$ is the degree of freedom of the Student’s t-distribution, \textit{i.e.}, temperature; $c$ is the number of clusters; $|m|$ is the number of heads; $\Gamma$ is an $[1 \times 1]$ convolutional operator; $\|$ is the concatenation operator; $\operatorname{GCont}$ is an auto-learned global graph content; $\operatorname{Gumbel-SoftMax}$ achieves soft sampling for neighborhood relationships to decrease the edge density.
\end{minipage} 
\caption{Summary of representative node clustering pooling methods in our framework\tnote{1}.}
\label{tab:node-clustering-pooling}
\end{table*}

% Minimize: $\|\boldsymbol{A} - \boldsymbol{U} \boldsymbol{H} \|_{F}$ 

%Auxiliary loss consists of link prediction objective $L_{\mathrm{LP}}=\|\boldsymbol{A}- \boldsymbol{C} \boldsymbol{C}^{T} \|_{F}$, where $\| \|_{F}$ denotes the Frobenius norm, and entropy regularization $L_{\mathrm{E}}=\frac{1}{c} \sum_{i=1}^{c} H\left(\boldsymbol{C}_{i}\right)$, where $H$ denotes the entropy function and $c$ is the number of clusters.
%%%%%%%%%%%%%%%%%%%%%%%%%%%%%%%%%%%%%%%%%%%%%%%%%%%%
%%%%%%%%%%%%%%%%%%%%%%%%%%%%%%%%%%%%%%%%%%%%%%%%%%%%

% \subsubsection{\textcircled{\raisebox{-0.9pt}{1}} Node Clustering Pooling}
\subsubsection{Node Clustering Pooling}
\label{sec:node-cluster}

Node clustering pooling considers graph pooling as a node clustering problem, which maps the nodes into a set of clusters. After that, the clusters are treated as new nodes of the new coarsened graph. To gain a better insight into node clustering pooling, we propose a universal and modularized framework to describe the process of node clustering pooling. Specifically, we deconstruct node clustering pooling into two disjoint modules: \textbf{1) Cluster Assignment Matrix (CAM) Generator.} Given an input graph, the CAM generator predicts the soft / hard assignment for each node. \textbf{2) Graph Coarsening.} With the assignment matrix, a new graph coarsened from the original one is obtained by learning a new feature matrix and a adjacency matrix. The process can be formulated as follows:
\begin{equation}
\begin{aligned}
&\underbrace{\boldsymbol{C}^{(l)}=\operatorname{CAM}(\boldsymbol{X}^{(l)}, \boldsymbol{A}^{(l)})}_{\text {\normalsize CAM Generator}} ;  \\
&\underbrace{\boldsymbol{X}^{(l+1)}, \boldsymbol{A}^{(l+1)}=\operatorname{COARSEN}(\boldsymbol{X}^{(l)}, \boldsymbol{A}^{(l)},\boldsymbol{C}^{(l)})}_{\text{\normalsize Graph Coarsening}}, 
\end{aligned} 
\end{equation}
where functions $\operatorname{CAM}$ and $\operatorname{COARSEN}$ are specially designed by each method, respectively. $\boldsymbol{C}^{(l)}\in \mathbb{R}^{ n_{l} \times n_{l+1}}$ indicates the learned cluster assignment matrix; $n_{l}$ is the number of nodes (or clusters) at layer $l$.

\begin{table*}[!t]
\renewcommand\arraystretch{2.0} % 行间距
\setlength\tabcolsep{30pt} % 列间距
\centering

\begin{threeparttable}[b]
%\footnotesize
%\scriptsize
%\tiny
\huge
\resizebox{0.9\textwidth}{!}{%
\begin{tabular}{@{}p{6.2cm}<{\centering}|c|c|c@{}}
\toprule
\multicolumn{1}{c}{\Huge \textbf{Models} }                                            & \multicolumn{1}{c}{\Huge \textbf{Score Generator} }                                                                                         & \multicolumn{1}{c}{\Huge  \textbf{Node Selector} }                                                                                                                  & \multicolumn{1}{c}{\Huge  \textbf{Graph Coarsening} }                                                                                                                                                                                                                                                                                                                                                                     \\ \midrule
\begin{tabular}[c]{@{}l@{}}\Huge TopKPool\tnote{[1]}\end{tabular} & $\boldsymbol{S}=\boldsymbol{X} \mathbf{p} /\left\|\mathbf{p}\right\|_{2}$                                                                                 & $\mathrm{idx}=\text{TOP}_{k}(\boldsymbol{S})$                                                                                              & $\begin{aligned}\boldsymbol{X}^{\prime}=\boldsymbol{X}_{\mathrm{idx}} \odot \sigma(\boldsymbol{S}_{\mathrm{idx}}); \quad \boldsymbol{A}^{\prime}=\boldsymbol{A}_{\mathrm{idx}, \mathrm{idx}}\end{aligned}$                                                                                                                                                                                                                                                                       \\ \midrule
\begin{tabular}[c]{@{}l@{}}\Huge SAGPool\tnote{[2]}\end{tabular}      & $\boldsymbol{S}= \operatorname{GNN}(\boldsymbol{X},\boldsymbol{A}) $                         & $\mathrm{idx}=\text{TOP}_{k}(\boldsymbol{S})$                                                                                             & $\begin{aligned} \boldsymbol{X}^{\prime}=\boldsymbol{X}_{\mathrm{idx}} \odot \boldsymbol{S}_{\mathrm{idx}}; \quad \boldsymbol{A}^{\prime}=\boldsymbol{A}_{\mathrm{idx}, \mathrm{idx}}\end{aligned}$                                                                                                                                                                                                                                                                       \\ \midrule
\begin{tabular}[c]{@{}l@{}}\Huge AttPool\tnote{[3]} \end{tabular}      & $\boldsymbol{S}= \operatorname{softmax}\left(\boldsymbol{X}\boldsymbol{W}\right)$                                                                                            & $\mathrm{idx}=\text{TOP}_{k}(\boldsymbol{S})$   & $\begin{aligned} \boldsymbol{X}^{\prime}= \boldsymbol{A}_{\mathrm{idx}}( \boldsymbol{X} \odot \boldsymbol{S}); \quad \boldsymbol{A}^{\prime}=\boldsymbol{A}_{\mathrm{idx}} \boldsymbol{A} \boldsymbol{A}_{\mathrm{idx}}^{T}   \end{aligned}$                                                                                                                                                                                                                                                                       \\ \midrule
\begin{tabular}[c]{@{}l@{}}\Huge ASAP\tnote{[4]}\end{tabular}            & $\boldsymbol{S} = LEConv(\boldsymbol{X^{c}},\boldsymbol{A})$                                                                                                           &  $\mathrm{idx}=\text{TOP}_{k}(\boldsymbol{S})$                                                                                              & $\begin{aligned} \boldsymbol{X}^{\prime} =\boldsymbol{X}^{c}_{\mathrm{idx}} \odot \boldsymbol{S}_{\mathrm{idx}}; \quad \boldsymbol{A}^{\prime}=\boldsymbol{A}_{\mathrm{idx}} \boldsymbol{A} \boldsymbol{A}_{\mathrm{idx}}^{T}   \end{aligned}$                                                                                                                                                                                                                                                               \\ \midrule
\begin{tabular}[c]{@{}l@{}}\Huge HGP-SL\tnote{[5]}\end{tabular}        & $\boldsymbol{S}=\left\|\left(\boldsymbol{I}-\boldsymbol{D}^{-1} \boldsymbol{A}\right) \mathbf{X}\right\|_{1}$                           & $\mathrm{idx}=\text{TOP}_{k}(\boldsymbol{S})$                                                                                              & $\left\{ \begin{aligned}&\boldsymbol{X}^{\prime}=\boldsymbol{X}_{\mathrm{idx}} \odot \boldsymbol{S}_{\mathrm{idx}}; \quad \widehat{\boldsymbol{A}}=\boldsymbol{A}_{\mathrm{idx}, \mathrm{idx}} \\& \boldsymbol{A}^{\prime}_{ij} = \operatorname{max}(\sigma(\overrightarrow{\mathbf{a}}[\boldsymbol{X}^{\prime}(i,:) \|\boldsymbol{X}^{\prime}(j,:)]^{\top})+\lambda \cdot \widehat{\boldsymbol{A}}_{ij} \end{aligned}\right.$                                                                                  \\ \midrule

\begin{tabular}[c]{@{}l@{}}\Huge VIPool\tnote{[6]}\end{tabular}      & $\left\{\begin{aligned}&\boldsymbol{P} = \frac{1}{t} \sum_{h=1}^{t} \left(\tilde{\boldsymbol{D}}^{-\frac{1}{2}} \tilde{\boldsymbol{A}} \tilde{\boldsymbol{D}}^{-\frac{1}{2}} \right)^{h} \boldsymbol{W}^{h} \text{MLP}(\boldsymbol{X}) \\& \boldsymbol{S} = \sigma(\text{MLP}(\text{MLP}(\boldsymbol{X}, \boldsymbol{P})) ) \end{aligned}\right.$ & $\mathrm{idx}=\text{TOP}_{k}(\boldsymbol{S})$                                                                                              & $\left\{ \begin{aligned} & \boldsymbol{X}^{\prime}=\boldsymbol{X}_{\mathrm{idx}} \odot \boldsymbol{S}_{\mathrm{idx}} \\& \boldsymbol{A}^{\prime}= softmax(\boldsymbol{A}_{\mathrm{idx}}) \boldsymbol{A} softmax( \boldsymbol{A}_{\mathrm{idx}})^{T} \end{aligned} \right.$                                                                                                                                                                                                                                                                                   \\ \midrule
\begin{tabular}[c]{@{}l@{}}\Huge RepPool\tnote{[7]}\end{tabular}     & $\boldsymbol{S}=\sigma\left(\boldsymbol{D}^{-1} \boldsymbol{A}  \boldsymbol{X} \mathbf{p} /\left\|\mathbf{p}\right\|_{2}\right)$                                                    & $\mathrm{idx}=\text{SEL}_{k}(\boldsymbol{S})$                                                                                              & $\left\{ \begin{aligned}&\boldsymbol{B}=\boldsymbol{X} \boldsymbol{W}_{b}(\boldsymbol{X}_{\mathrm{idx}})^{T}  \\ &\boldsymbol{X}^{\prime}=(\operatorname{softmax}(\boldsymbol{B} \odot \boldsymbol{M}))^{T} (\boldsymbol{X} \odot\boldsymbol{S})\\ &\boldsymbol{A}^{\prime}=(\operatorname{softmax}(\boldsymbol{B} \odot \boldsymbol{M}))^{T} \boldsymbol{A}(\operatorname{softmax}(\boldsymbol{B} \odot \boldsymbol{M})) \end{aligned} \right.$ \\ \midrule

%\begin{tabular}[c]{@{}l@{}}UGPool~\shortcite{uniform-pooling}\end{tabular}     & $\boldsymbol{S}=\sigma(\boldsymbol{X} \mathbf{p} /\left\|\mathbf{p}\right\|_{2}) $                                                    & $\mathrm{idx}=\text{1DPool}(\text{rank}(\boldsymbol{S}))$                                                                                              &  $\begin{aligned} \boldsymbol{X}^{\prime}=\boldsymbol{X}_{\mathrm{idx}} \odot \boldsymbol{S}_{\mathrm{idx}}; \quad \boldsymbol{A}^{\prime}=\boldsymbol{A}_{\mathrm{idx}, \mathrm{idx}} + \boldsymbol{A}_{\mathrm{idx}, \mathrm{idx}}^{2} \end{aligned}$  \\ %\midrule

\begin{tabular}[c]{@{}l@{}}\Huge GSAPool\tnote{[8]}\end{tabular}      & $\left\{\begin{aligned}&\boldsymbol{S}_{1}=\operatorname{GNN}( \boldsymbol{X},  \boldsymbol{A})   \\& \boldsymbol{S}_{2} = \sigma(\operatorname{MLP}(\boldsymbol{X})) \\& \boldsymbol{S} = \alpha \boldsymbol{S}_{1} + (1- \alpha)\boldsymbol{S}_{2}\end{aligned}\right.$ & $\mathrm{idx}=\text{TOP}_{k}(\boldsymbol{S})$                                                                                              & $ \boldsymbol{X}^{\prime}=(\boldsymbol{A}\boldsymbol{X}\boldsymbol{W})_{\mathrm{idx}} \odot \boldsymbol{S}_{\mathrm{idx}}; \quad \boldsymbol{A}^{\prime}=\boldsymbol{A}_{\mathrm{idx}, \mathrm{idx}}$                                                                                                                                                                                                                                                                                    \\ \midrule

\begin{tabular}[c]{@{}l@{}}\Huge PANPool\tnote{[9]}\end{tabular}      & $\left\{\begin{aligned}& \boldsymbol{Z}_{i}=\sum_{n=0}^{L} e^{-\frac{E(n)}{T}} \sum_{j=1}^{N} g(i, j ; n) \\& \boldsymbol{M}=\boldsymbol{Z}^{-1} \sum_{n=0}^{L} e^{-\frac{E(n)}{T}} \boldsymbol{A}^{n} \\& \boldsymbol{S} =\boldsymbol{X} \mathbf{p} +\beta \operatorname{diag}(\boldsymbol{M}) \end{aligned}\right.$ 
     & $\mathrm{idx}=\text{TOP}_{k}(\boldsymbol{S})$                                                                                    & $\begin{aligned}\boldsymbol{X}^{\prime}=\boldsymbol{X}_{\mathrm{idx}} \odot \sigma(\boldsymbol{S}_{\mathrm{idx}}); \quad \boldsymbol{A}^{\prime}=\boldsymbol{A}_{\mathrm{idx}, \mathrm{idx}}\end{aligned}$                                                                                                                                                                                                                                                                         \\ \midrule

\begin{tabular}[c]{@{}l@{}}\Huge CGIPool\tnote{[10]}\end{tabular}      & $\left\{\begin{aligned}&\boldsymbol{S}_{r}=\operatorname{GNN}_{r}( \boldsymbol{X},  \boldsymbol{A}) \\& \boldsymbol{S}_{f}=\operatorname{GNN}_{f}( \boldsymbol{X},  \boldsymbol{A}) \\& \boldsymbol{S} = \sigma( \boldsymbol{S}_{r} - \boldsymbol{S}_{f}) \end{aligned}\right.$ & $\mathrm{idx}=\text{TOP}_{k}(\boldsymbol{S})$                                                                                              & $\begin{aligned} \boldsymbol{X}^{\prime}=\boldsymbol{X}_{\mathrm{idx}} \odot \boldsymbol{S}_{\mathrm{idx}}; \quad \boldsymbol{A}^{\prime}=\boldsymbol{A}_{\mathrm{idx}, \mathrm{idx}}\end{aligned} $                                                                                                                                                                                                                                                                                   \\ \midrule

\begin{tabular}[c]{@{}l@{}}\Huge TAPool\tnote{[11]}\end{tabular}      & $\left\{\begin{aligned}&\boldsymbol{S}_{l}=\operatorname{softmax}\left( \frac{1}{n} ((\boldsymbol{X} \boldsymbol{X}^{T}) \odot (\widetilde{\boldsymbol{D}}^{-1} \widetilde{\boldsymbol{A}})) \boldsymbol{1}_{n} \right)  \\& \boldsymbol{S}_{g}= \operatorname{softmax} \left(\widetilde{\boldsymbol{D}}^{-1} \widetilde{\boldsymbol{A}} \boldsymbol{X} \mathbf{p} \right)  \\& \boldsymbol{S} =  \boldsymbol{S}_{l} + \boldsymbol{S}_{g} \end{aligned}\right.$ & $\mathrm{idx}=\text{TOP}_{k}(\boldsymbol{S})$                                                                                              & $\begin{aligned} \boldsymbol{X}^{\prime}=\boldsymbol{X}_{\mathrm{idx}} \odot \boldsymbol{S}_{\mathrm{idx}}; \quad \boldsymbol{A}^{\prime}=\boldsymbol{A}_{\mathrm{idx}, \mathrm{idx}}\end{aligned} $                                                                                                                                                                                                                                                                                   \\ \midrule

\begin{tabular}[c]{@{}l@{}}\Huge IPool\tnote{[12]}\end{tabular}          & $\boldsymbol{S}=\left\|\left(\boldsymbol{I}-\frac{1}{t} \sum_{h=1}^{t} (\bar{\boldsymbol{D}}^{h})^{-1} \bar{\boldsymbol{A}}^{h} \right) \boldsymbol{X}\right\|_{2} $                                                   & $\mathrm{idx}=\text{TOP}_{k}(\boldsymbol{S})$  & $ \left\{ \begin{aligned}&\boldsymbol{X}^{\prime}=\boldsymbol{X}_{\mathrm{idx}}, \\&\boldsymbol{A}^{\prime}_{i j}=\lambda\left(\boldsymbol{A}+\boldsymbol{I}\right)_{\mathrm{idx}[i], \mathrm{idx}[j]} + (1- \lambda) \boldsymbol{O}_{ij}  \end{aligned}\right. $                                                                                                                                                                                                                          \\ \bottomrule
\end{tabular}%
}
%\begin{tablenotes}
%  \tiny 
%%  \item[1] This manuscript only presents the global attention mechanism to generate scores, and there is  also a local attention mechanism introduced in ~\cite{attpool}.  
%  \item[1] This manuscript only presents the greedy  IPool strategy in this table, and there is  also a local IPool strategy  introduced in ~\cite{ipool}. 
%  \end{tablenotes}
  \end{threeparttable}
  \begin{minipage}{0.9\linewidth} \scriptsize %\footnotesize %\scriptsize %\tiny
 $^{[1]}$\cite{graph-u-net};  $^{[2]}$\cite{sagpool};   $^{[3]}$\cite{attpool};   $^{[4]}$~\cite{asap}; $^{[5]}$\cite{hgp-sl};   $^{[6]}$\cite{vip-pool};   $^{[7]}$\cite{rep-pool};   $^{[8]}$\cite{gsapool};   $^{[9]}$\cite{path-pooling};  $^{[10]}$\cite{cgi-pool};  $^{[11]}$\cite{TAPool};  $^{[12]}$\cite{ipool}.
 
\textbf{Notations:} $\boldsymbol{X}^{\prime} \in \mathbb{R}^{ k \times d}$ and  $\boldsymbol{A}^{\prime} \in \{0, 1\}^{ k \times k}$ are the adjacency matrix and feature matrix for the new graph, respectively; $\boldsymbol{D} \in \mathbb{R}^{ n \times n} $ is the degree matrix of $\boldsymbol{A}$; $\bar{\boldsymbol{A}}^{h} \in \mathbb{R}^{ n \times n}$ is the matrix where diagonal values corresponding to the $h$-hop circles have been removed; $\bar{\boldsymbol{D}}^{h} \in \mathbb{R}^{ n \times n}$ is the corresponding degree matrix of $\bar{\boldsymbol{A}}^{h}$; $\boldsymbol{W} \in \mathbb{R}^{ d \times 1}$; and  $\boldsymbol{W}_{b} \in \mathbb{R}^{ d \times d}$ are the learnable weight matrices;  $\mathbf{p} \in \mathbb{R}^{d}$ and $\mathbf{a} \in \mathbb{R}^{ 1 \times 2d}$ are the trainable projection vectors; $\boldsymbol{I}$ is the identity matrix;  $\lambda $ is a trade-off parameter;  $\alpha$ is a user-defined hyperparameter; $\mathbf{1}_{n} \in \mathbb{R}^{n} $ is a vector with all elements being 1; $\boldsymbol{M}  \in \mathbb{R}^{ n \times k} $ is a masking matrix; $\boldsymbol{O}$ is the matrix used for measuring the overlap between node neighbors; $\sigma$ is the activation function (\textit{e.g.}, $\operatorname{tanh}$);  $\odot$ is the broadcasted elementwise product; $\text{MLP}$ is a multi-layer perceptron; $\text{SEL}_{k}$ is the algorithm used for selecting nodes one by one.
\end{minipage}
\caption{Summary of representative node drop pooling methods in our framework\tnote{1}.}
\label{tab:node-drop-pooling}
\end{table*}

Accordingly, we show how existing node clustering pooling methods fit into our proposed framework, and select seven typical methods presented in Table~\ref{tab:node-clustering-pooling}. We observe that these methods, with the same coarsening module choice, mainly differ in the way CAM is generated.  \textbf{1) CAM Generator.} Node clustering methods generate CAM from different perspectives. Specifically, DiffPool~\cite{diffpool} directly employs GNN models, and StructPool~\cite{structpool} extends DiffPool by explicitly capturing high-order structural relationships;  LaPool~\cite{lap-pool} and MinCutPool~\cite{mincut} both design the generator from the perspective of spectral clustering;  MemPool~\cite{mem-pool} introduces a clustering-friendly distribution to generate the cluster matrix.
\textbf{2) Graph Coarsening.} Most node clustering methods adopt nearly the same coarsening strategy: the pooled node representations, $\boldsymbol{X}^{(l+1)} ={\boldsymbol{C}^{(l)}}^{T} \boldsymbol{X}^{(l)} \in \mathbb{R}^{ n_{l+1} \times d}$, obtained by the sum of representations of the nodes in each cluster and weighted by the cluster assignment scores;  the coarsened adjacency matrix, $\boldsymbol{A}^{(l+1)} = {\boldsymbol{C}^{(l)}}^{T} \boldsymbol{A}^{(l)} \boldsymbol{C}^{(l)} \in \mathbb{R}^{ n_{l+1} \times n_{l+1}} $, which indicates the connectivity strength between different clusters, is obtained by the weighted sum of edges between clusters.

Due to space limitation, many other node clustering pooling methods~\cite{eigenpool,muchGNN,deep-graph-mapper,mem-pool,haar-pooling,HIBpool,capsule-pooling,mxpool,LCP,smip} are not presented in Table~\ref{tab:node-clustering-pooling}. Despite substantial improvements achieved on several graph-level tasks (\textit{e.g.}, graph classification), the above methods suffer from the limitation of the time and storage complexity. This is due to the computation of a dense cluster assignment matrix, which typically requires $\mathcal{O}\left(n^2\right)$ space complexity, as noted in~\cite{gmt}.  Besides, as discussed in the recent work~\cite{rethink-pooling}, clustering-enforcing regularization usually has little effect.

% \subsubsection{\textcircled{\raisebox{-0.9pt}{2}} Node Drop Pooling}

\subsubsection{Node Drop Pooling}
Node drop pooling exploits learnable scoring functions to delete nodes with comparatively lower significance scores. For a thorough analysis of node drop pooling, we propose a universal and modularized framework, which consists of three disjoint modules: \textbf{1) Score Generator.} Given an input graph, the score generator calculates significance scores for each node. \textbf{2) Node Selector.} Node selector selects the nodes with top-k significance scores.  \textbf{3) Graph Coarsening.} With the selected nodes, a new graph coarsened from the original one is obtained by learning a new feature matrix and a adjacency matrix. The process can be formulated as follows:
\begin{equation}
\begin{aligned}
&\underbrace{\boldsymbol{S}^{(l)}=\operatorname{SCORE}(\boldsymbol{X}^{(l)}, \boldsymbol{A}^{(l)})}_{\text {\normalsize Score Generator}} ;   \quad \underbrace{\mathrm{idx}^{(l+1)}=\operatorname{TOP}_{k}(\boldsymbol{S}^{(l)})}_{\text{\normalsize Node Selector}} ;  \\
&\underbrace{\boldsymbol{X}^{(l+1)}, \boldsymbol{A}^{(l+1)}=\operatorname{COARSEN}(\boldsymbol{X}^{(l)}, \boldsymbol{A}^{(l)},\boldsymbol{S}^{(l)},\mathrm{idx}^{(l+1)})}_{\text{\normalsize Graph Coarsening}}, 
\end{aligned} 
\end{equation}
where functions $\operatorname{SCORE}$, $\operatorname{TOP}_{k}$, and $\operatorname{COARSEN}$ are specially designed by each method for score generator, node selector, and graph coarsening, respectively. $\boldsymbol{S}^{(l)}\in \mathbb{R}^{ n \times 1}$ indicates the significance scores; $\operatorname{TOP}_{k}$ ranks values and returns the indices of the largest $k$ values in $\boldsymbol{S}^{(l)}$;  $\mathrm{idx}^{(l+1)}$ indicates the reserved node indexes for the new graph.

Accordingly, we present how the nine typical node drop pooling methods fit into our proposed framework in Table~\ref{tab:node-drop-pooling}. Intuitively, methods tend to design more sophisticated score generators and more reasonable graph coarsenings to select more representative nodes and retain more important structural information, respectively, thus alleviating the problem of information loss. \textbf{1) Score Generator.} Different from TopKPool~\cite{graph-u-net}, SAGPool~\cite{sagpool}, and HGP-SL~\cite{hgp-sl}, which predict scores from a single view, GSAPool~\cite{gsapool} and TAPool~\cite{TAPool} generate scores from two different views, \textit{i.e.}, local and global views. \textbf{2) Node Selector.} Most methods simply adopt $\operatorname{TOP}_{k}$ as a selector, and only a few works~\cite{rep-pool,uniform-pooling} design different selectors. \textbf{3) Graph Coarsening.} Instead of directly obtaining the coarsened graph formed by the selected nodes, such as TopKPool~\cite{graph-u-net}, SAGPool~\cite{sagpool}, and TAPool~\cite{TAPool}, RepPool~\cite{rep-pool}, GSAPool~\cite{gsapool}, and IPool~\cite{ipool} utilize both the selected nodes and non-selected nodes to maintain more structural and feature information in a graph.

Due to space limitation, many other node drop pooling methods~\cite{topkpool,understand-attention,asap,path-pooling,NDPool,lookhops,vip-pool,MVpool,commpool,liu2022on,MGAP} are not presented in Table~\ref{tab:node-drop-pooling}. Though more efficient and more applicable to large-scale graph datasets~\cite{topkpool} than node clustering pooling methods, node drop pooling methods suffer from inevitable information loss~\cite{ipool,gmt,liu2022on}.
%
%Later, some popular graph clustering algorithms~\cite{graclus} have been exploited to perform pooling in GNN architectures~\cite{CNN-with-fast}. 

% \subsubsection{\textcircled{\raisebox{-0.9pt}{3}} Other Pooling}
\subsubsection{Other Pooling}
Apart from node drop  and node clustering pooling methods, there also exist some other graph pooling methods. For example, EdgePool~\cite{edgepool} and HyperDrop~\cite{hyper-pool} pool the input graph from the edge view, which maintains the connectivity of the input graph through edge contractions based on an adaptive edge scoring design; MuchPool~\cite{much-pool} combines node clustering pooling and node drop pooling to capture different characteristics of a graph; PAS~\cite{pooling-search} proposes to search for adaptive pooling architectures by neural architecture search.

%Co-Pooling~\cite{edge-view}

\begin{table}[!t]
\renewcommand\arraystretch{1.0} % 行间距
\setlength\tabcolsep{3pt} % 列间距
\centering
\tiny
\begin{threeparttable}[b]
\resizebox{0.48\textwidth}{!}{%
\begin{tabular}{@{}ccccccc@{}}
\toprule
                          \textbf{ Datasets}  &  \textbf{Category}           & \textbf{ \# graphs} & \textbf{\# classes} & \textbf{Avg.$|\mathcal{V}|$}         & \textbf{Avg.$|\mathcal{E}|$}   & \textbf{ Task}       \\ \midrule
                         
%%%%%%%%%%%%%%%%%%%%%%%%%%%%%%%%%%%%%%%%%%%%%%%%%%%%%%%%%%%%%%%%%%%%%%%%%%%
%%%%%%%%%%%%%%%%%%%%%%%%%%%%%%%%%%%%%%%%%%%%%%%%%%%%%%%%%%%%%%%%%%%%%%%%%%%
%%%%%%%%%%%%%%%%%%%%%%%%%%%%%%%%%%%%%%%%%%%%%%%%%%%%%%%%%%%%%%%%%%%%%%%%%%%

\rowcolor{gray!20} \multicolumn{7}{c}{TUDataset\tnote{[1]}} \\
\midrule

 D\&D  & Protein         & 1,178     & 2          & 284.32               & 715.66    & Cla.              \\
                               PROTEINS & Protein    & 1,113     & 2          & 39.06                & 72.82    & Cla.               \\
                               ENZYMES  & Protein    & 600       & 6          & 32.63                & 124.20     & Cla.             \\ \midrule
                               NCI1    & Molecule     & 4,110     & 2          & 29.87                & 32.30     & Cla.             \\
                               NCI109  & Molecule     & 4,127     & 2          & 29.68                & 32.13       & Cla.            \\
                        		   MUTAG  & Molecule     & 188       & 2          & 17.93                & 19.79      & Cla.              \\
                        		   PTC\_MR & Molecule      & 344       & 2          & 14.30                & 14.69      & Cla.             \\
                               MUTAGENICITY & Molecule & 4,337     & 2          & 30.32                & 30.77     & Cla.              \\
                               
                               FRANKENSTEIN & Molecule & 4,337     & 2          & 16.90                & 17.88     & Cla.              \\ \midrule
                               REDDIT-BINARY &Social     & 2,000     & 2          & 429.63               & 497.75      & Cla.            \\
                               REDDIT-M5K &Social   & 4,999     & 5          & 508.52               & 594.87     & Cla.             \\
                               REDDIT-M12K &Social     & 11,929     & 11          & 391.41              & 456.89      & Cla.            \\
                               IMDB-BINARY &Social      & 1,000     & 2          & 19.77                & 96.53    & Cla.            \\
                               IMDB-MULTI  &Social    & 1,500     & 3          & 13.00                & 65.94      & Cla.            \\
                               COLLAB   &Social  & 5,000     & 3          & 74.49                & 2457.78      & Cla.  \\     \midrule     
                               QM9 & Molecule          & 133,885   & --          & 18.03               & 18.63   & Reg.                   \\
                               ZINC & Molecule       & 249,456     & --         & 23.14               & 24.91     & Rec.\\ \midrule

%%%%%%%%%%%%%%%%%%%%%%%%%%%%%%%%%%%%%%%%%%%%%%%%%%%%%%%%%%%%%%%%%%%%%%%%%%%
%%%%%%%%%%%%%%%%%%%%%%%%%%%%%%%%%%%%%%%%%%%%%%%%%%%%%%%%%%%%%%%%%%%%%%%%%%%
%%%%%%%%%%%%%%%%%%%%%%%%%%%%%%%%%%%%%%%%%%%%%%%%%%%%%%%%%%%%%%%%%%%%%%%%%%%
\rowcolor{gray!20} \multicolumn{7}{c}{Open Graph Benchmark\tnote{[2]}} \\
\midrule
 HIV &Molecule       & 41,127    & 2          & 25.51                & 27.52    & Cla.                    \\
                               TOX21   &Molecule     & 7,831     & 12         & 18.57                & 19.3    & Cla.               \\
                              TOXCAST  &Molecule     & 8,576     & 617        & 18.78                & 19.3      & Cla.              \\
                              BBBP    &Molecule      & 2,039     & 2          & 24.06                & 26.0    & Cla.             \\ \midrule

%%%%%%%%%%%%%%%%%%%%%%%%%%%%%%%%%%%%%%%%%%%%%%%%%%%%%%%%%%%%%%%%%%%%%%%%%%%
%%%%%%%%%%%%%%%%%%%%%%%%%%%%%%%%%%%%%%%%%%%%%%%%%%%%%%%%%%%%%%%%%%%%%%%%%%%
%%%%%%%%%%%%%%%%%%%%%%%%%%%%%%%%%%%%%%%%%%%%%%%%%%%%%%%%%%%%%%%%%%%%%%%%%%%
\rowcolor{gray!20} \multicolumn{7}{c}{MoleculeNet\tnote{[3]}} \\
\midrule

QM7 & Molecule         & 7,165   & 1          & 61.31                & 91.03     & Reg.                   \\
QM8 & Molecule          & 21,786   & 12          & 61.31                & 91.03     & Reg.                   \\
						 ESOl & Molecule          & 1,128   & 1          & 61.31                & 91.03   & Reg.                   \\
                               FREESOLV  & Molecule        & 643     & 1         & 20.85                & 32.74     & Reg.  
     \\
                               LIPOPHILICITY  & Molecule       & 4,200     & 1         & 20.85                & 32.74     & Reg.                 \\ \midrule
%%%%%%%%%%%%%%%%%%%%%%%%%%%%%%%%%%%%%%%%%%%%%%%%%%%%%%%%%%%%%%%%%%%%%%%%%%%
%%%%%%%%%%%%%%%%%%%%%%%%%%%%%%%%%%%%%%%%%%%%%%%%%%%%%%%%%%%%%%%%%%%%%%%%%%%
\rowcolor{gray!20} \multicolumn{7}{c}{Synthetic Generation\tnote{[4]}} \\
\midrule
 COLORS-3 & Synthetic         & 5,500   & 5          & 61.31                & 91.03    & Cla.                   \\
                               TRIANGLES  & Synthetic     & 45,000     & 10         & 20.85                & 32.74    & Cla.                \\ \midrule

%%%%%%%%%%%%%%%%%%%%%%%%%%%%%%%%%%%%%%%%%%%%%%%%%%%%%%%%%%%%%%%%%%%%%%%%%%%
%%%%%%%%%%%%%%%%%%%%%%%%%%%%%%%%%%%%%%%%%%%%%%%%%%%%%%%%%%%%%%%%%%%%%%%%%%%
\rowcolor{gray!20} \multicolumn{7}{c}{Computer Vision\tnote{[5]}} \\
\midrule
MNIST & Image         & 60,000   & 10          & 70                & 91.03  & Cla.                   \\
                               CIFAR10  & Image      & 70,000     & 10         & 117                & 32.74    & Cla.             \\%\midrule
                              
%\multirow{2}{*}{\textbf{fMRI}\tnote{5}}  & ABIDE & Image         & 2,614   & 2          & 808                & 8080 & -- & Classification                    \\
%                              & HCP  & Image      & 4,233     & 2         & 808                & 8080  & -- & Classification               \\
                              
\bottomrule
\end{tabular}%
}
%\begin{tablenotes}
%  \small 
%  \item[1] ~\url{https://chrsmrrs.github.io/datasets/docs/datasets/} 
%  \item[2] ~\url{https://ogb.stanford.edu/docs/graphprop/} 
%  \item[3] ~\url{https://github.com/bknyaz/graph_attention_pool/tree/master/data}
%  \item[4] ~\url{https://github.com/graphdeeplearning/benchmarking-gnns} 
%  \end{tablenotes}
  \end{threeparttable}

  \begin{minipage}{\linewidth} \scriptsize %\footnotesize \tiny
  Cla., Reg., and Rec. refer to graph classification, graph regression, and  graph reconstruction, respectively. [1]~\url{https://chrsmrrs.github.io/datasets/docs/datasets/}; [2]~\url{https://ogb.stanford.edu/docs/graphprop/}; [3]~\url{https://moleculenet.org/}; [4]~\url{https://github.com/bknyaz/graph_attention_pool/tree/master/data};  [5]~\url{https://github.com/graphdeeplearning/benchmarking-gnns}
\end{minipage}
\caption{A list of commonly used and publicly accessible datasets.}
\label{tab:dataset}
\end{table}

\begin{figure}[!t] % !htb
\begin{center}

\includegraphics[width=1.0\linewidth]{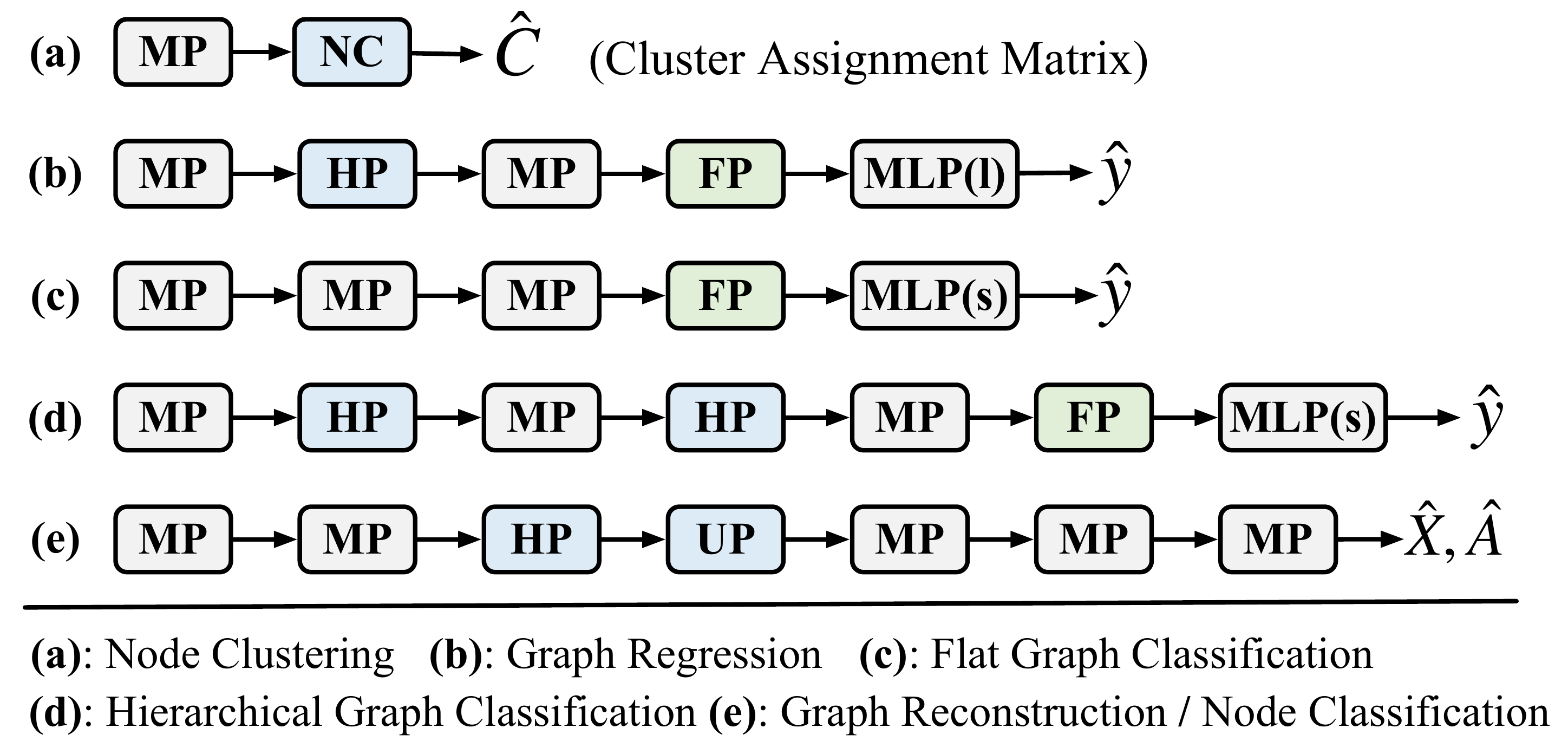}
\end{center}
\caption{Illustration of model architectures for different tasks.}
\label{fig:model-arch}
\end{figure}

%%%%%%%%%%%%%%%%%%%%%%%%%%%%%%%%%%%%%%%%%%%%%%%%%%%%%%%%%%%%%%%%%%%%%
%%%%%%%%%%%%%%%%%%%%%%%%%%%%%%%%%%%%%%%%%%%%%%%%%%%%%%%%%%%%%%%%%%%%%
%%%%%%%%%%%%%%%%%%%%%%%%%%%%%%%%%%%%%%%%%%%%%%%%%%%%%%%%%%%%%%%%%%%%%

\begin{table*}[!t]
\centering

\begin{threeparttable}[b]
%\footnotesize
%\scriptsize
\renewcommand\arraystretch{1.32} % 行间距
\resizebox{\textwidth}{!}{%
\begin{tabular}{@{}p{2.8cm}<{\centering}|l|l|c|c@{}}
\toprule
\multicolumn{1}{c}{\large \textbf{Method} }                                          & \multicolumn{1}{c}{\large  \textbf{Task} }                                                                                         & \multicolumn{1}{c}{\large  \textbf{Dataset} }                                                                                                                  
& \multicolumn{1}{c}{\large  \textbf{Venue} } 
& \multicolumn{1}{c}{\large  \textbf{Code Link} }                                                                                                                                                                                                                                                                                                                                                                     \\ %\midrule

\midrule
\rowcolor{gray!20} \multicolumn{5}{c}{ \Large Flat Pooling Methods} \\
\midrule

\begin{tabular}[c]{@{}l@{}}SortPool\tnote{[1]$\spadesuit$$\clubsuit$}\end{tabular}      & Graph Classification                        & D\&D, PROTEINS, NCI1, MUTAG, PTC, COLLAB, IMDB-B (M)                                                                        & AAAI'2018 &  { \url{https://shorturl.at/joAJ9} }                                                                                                                                                                                                                                                                     \\ \midrule

\begin{tabular}[c]{@{}l@{}}SOPool\tnote{[2]}\end{tabular}      & Graph Classification                                                                                & PROTEINS, NCI1, MUTAG, PTC, IMDB-B (M), COLLAB, RDT-B,RDT-M5K & TPAMI'2020                                                                                           & --                                                                                                                                                                                                                                                                                              \\ \midrule

\multirow{3}{*}{GMT\tnote{[3]$\spadesuit$}}  & Graph Classification &  D\&D, PROTEINS, MUTAG, IMDB-B (M), COLLAB, HIV, TOX21, TOXC., BBBP&   \multirow{3}{*}{ICLR'2021}      &   \multirow{3}{*}{  \url{https://shorturl.at/iorN3}}                   \\
                             & Graph Reconstruction    & ZINC,  Synthetic (Ring and Grid Graphs) & & \\
                             & Graph Generation    & QM9 & &        \\ \midrule

\begin{tabular}[c]{@{}l@{}}DKEPool\tnote{[4]}\end{tabular}      & Graph Classification                                                                                & PROTEINS, NCI1, MUTAG, PTC, IMDB-B (M), HIV, BBBP & TKDE'2022                                                                                           &  \url{https://shorturl.at/drz12}                                                                                                                                                                                                                                                                                             \\ %\midrule
%%%%%%%%%%%%%%%%%%%%%%%%%%%%%%%%%%%%%%%%%%%%%%%%%%%%%%%%%%%%%%%%%%%%%%%%%%%%%%
%%%%%%%%%%%%%%%%%%%%%%%%%%%%%%%%%%%%%%%%%%%%%%%%%%%%%%%%%%%%%%%%%%%%%%%%%%%%%%
%%%%%%%%%%%%%%%%%%%%%%%%%%%%%%%%%%%%%%%%%%%%%%%%%%%%%%%%%%%%%%%%%%%%%%%%%%%%%%
\midrule
\rowcolor{gray!20} \multicolumn{5}{c}{  \Large Node Clustering Pooling Methods} \\
\midrule

\begin{tabular}[c]{@{}l@{}}DiffPool\tnote{[5]$\spadesuit$$\clubsuit$}\end{tabular}     & Graph Classification                        & D\&D, PROTEINS, ENZYMES, COLLAB, RDT-M12K                                                                        & NeurIPS'2018 &  {  \url{https://shorturl.at/aflB3} }                                                                                                                                                                                                                                                                     \\ \midrule

% \begin{tabular}[c]{@{}l@{}}EigenPool\tnote{[6]} \end{tabular}   & Graph Classification                                                                                & D\&D, PROTEINS,ENZYMES, NCI1, NCI109, MUTAG & KDD'2019                                                                                           &{  \url{https://shorturl.at/iqIS7} }                                                                                                                                                                                                                                                                               \\ \midrule

\multirow{3}{*}{MinCutPool\tnote{[6]$\spadesuit$$\clubsuit$}}   & Graph Classification &  D\&D, PROTEINS, ENZYMES, IMDB-B (M), COLLAB &   \multirow{3}{*}{ICML'2020}      &   \multirow{3}{*}{ \url{https://shorturl.at/zIOY9}}                   \\
                             & Node Clustering    & Cora, Citeseer, Pubmed & & \\
                             & Graph Regression    & QM9  & &       \\ \midrule

% \multirow{2}{*}{HaarPool\tnote{[8]}} & Graph Classification &   PROTEINS, NCI1, NCI109, MUTAG, MUTAGEN., TRIANGLES &   \multirow{2}{*}{ICML'2020}      &   \multirow{2}{*}{ \url{https://shorturl.at/CY239}}                   \\
%                              & Graph Regression    & QM9          \\ \midrule

\multirow{2}{*}{MemPool\tnote{[7]}}   & Graph Classification &   D\&D, PROTEINS, ENZYMES, COLLAB, RDT-B &   \multirow{2}{*}{ICLR'2020}      &   \multirow{2}{*}{ \url{https://shorturl.at/dqxY4}}                   \\
                             & Graph Regression    & ESOL, Lipophilicity  & &        \\ \midrule

\begin{tabular}[c]{@{}l@{}}StructPool\tnote{[8]}\end{tabular}     & Graph Classification                                                                                & PROTEINS, ENZYMES,MUTAG, PTC, IMDB-B (M),COLLAB & ICLR'2020                                                                                           & { \url{https://shorturl.at/fiJ01} }                                                                                                                                                                                                                                                                                         \\ \midrule

\multirow{2}{*}{HoscPool\tnote{[9]}}  & Graph Classification &  D\&D, PROTEINS, NCI1, MUTAGEN., RDT-B, COX2-MD, ER-MD &   \multirow{2}{*}{CIKM'2022}      &   \multirow{2}{*}{\url{https://shorturl.at/abR26}}                   \\
                             & Node Clustering    & Cora, Pubmed, CS, Photo, PC, Polblogs, Eu-email, Synthetic           \\ \midrule

\multirow{3}{*}{SEP\tnote{[10]}}  & Graph Classification &  D\&D, PROTEINS, MUTAG, NCI1, IMDB-B (M), COLLAB&   \multirow{3}{*}{ICML'2022}      &   \multirow{3}{*}{  \url{https://shorturl.at/ovDLT}}                   \\
                             & Graph Reconstruction    & Synthetic (Ring and Grid Graphs) & &  \\
                             & Node Classification   & Cora, Citeseer, Pubmed  & &       \\ %\midrule
%%%%%%%%%%%%%%%%%%%%%%%%%%%%%%%%%%%%%%%%%%%%%%%%%%%%%%%%%%%%%%%%%%%%%%%%%%%%%%
%%%%%%%%%%%%%%%%%%%%%%%%%%%%%%%%%%%%%%%%%%%%%%%%%%%%%%%%%%%%%%%%%%%%%%%%%%%%%%
%%%%%%%%%%%%%%%%%%%%%%%%%%%%%%%%%%%%%%%%%%%%%%%%%%%%%%%%%%%%%%%%%%%%%%%%%%%%%%
\midrule
\rowcolor{gray!20} \multicolumn{5}{c}{  \Large Node Drop Pooling Methods} \\
\midrule

\
\multirow{2}{*}{TopKPool\tnote{[11]$\spadesuit$$\clubsuit$} }   
    & Graph Classification & D\&D, PROTEINS, COLLAB  &   \multirow{2}{*}{ICML'2019}      &   \multirow{2}{*}{ \url{https://shorturl.at/cjlnr} }                   \\ &
                               Node Classification    & Cora, Citeseer, Pubmed  & &         \\ \midrule

\begin{tabular}[c]{@{}l@{}}SAGPool\tnote{[12]$\spadesuit$$\clubsuit$}\end{tabular}      & Graph Classification                        & D\&D, PROTEINS, NCI1, NCI109, FRANK.                                                                         & ICML'2019 &  { \url{https://shorturl.at/bEJNQ}}                                                                                                                                                                                                                                                                      \\ \midrule

% \begin{tabular}[c]{@{}l@{}}AttPool\tnote{[15]$\spadesuit$} \end{tabular}  & Graph Classification                                                                                & D\&D, PROTEINS,NCI1, COLLAB, RDT-B,RDT-M12K & ICCV'2019                                                                                           & { \url{https://shorturl.at/HJMX1}}                                                                                                                                                                                                                                                                                \\ \midrule

%\begin{tabular}[c]{@{}l@{}}EdgePool~\cite{edgepool}\tnote{$\spadesuit$} \end{tabular} & OT  & Graph Classification                                                                                & PROTEINS, COLLAB, RDT-B, RDT-M12K & ICML-W'2019                                                                                           & {\footnotesize \url{https://github.com/pyg-team/pytorch_geometric}}                                                                                                                                                                                                                                                                                \\ \midrule

\begin{tabular}[c]{@{}l@{}}ASAP\tnote{[13]$\spadesuit$}\end{tabular}        & Graph Classification                                                                                & D\&D, PROTEINS, ENZYMES, NCI1, NCI109, MUTAGEN. & AAAI'2020                                                                                           & { \url{https://shorturl.at/depuz}}                                                                                                                                                                                                                                                                           \\ \midrule
% \begin{tabular}[c]{@{}l@{}}HGP-SL\tnote{[17]}\end{tabular}        & Graph Classification                                                                                & D\&D, PROTEINS, NCI1, NCI109, FRANK. & AAAI'2020                                                                                           & { \url{https://shorturl.at/qyEJ6}}                                                \\ \midrule

\multirow{2}{*}{VIPool\tnote{[14]}}  & Graph Classification &  D\&D, PROTEINS, ENZYMES, IMDB-B (M), COLLAB &   \multirow{2}{*}{NeurIPS'2020}      &   \multirow{2}{*}{\url{https://shorturl.at/luvz1}}                   \\
                             & Node Classification    & Cora, Citeseer, Pubmed   & &        \\ \midrule

% \begin{tabular}[c]{@{}l@{}}RepPool\tnote{[19]}\end{tabular}    & Graph Classification                                                                                & D\&D, PROTEINS, NCI1, NCI109, MUTAG, PTC, IMDB-B & ICDM'2020                                                                                           & {\url{https://shorturl.at/KZ457} }                    
%  \\ \midrule

\begin{tabular}[c]{@{}l@{}}GSAPool\tnote{[15]}\end{tabular}     & Graph Classification                                                                                & D\&D, NCI1, NCI109, MUTAGEN. & WWW'2020                                                                                           & {  \url{https://shorturl.at/ciS02} }                                                                                                                                                                                                                                                                                         \\ \midrule

\begin{tabular}[c]{@{}l@{}}PANPool\tnote{[16]$\spadesuit$}\end{tabular}     & Graph Classification                                                                                & PROTEINS, PROTEINS-FULL, NCI1, MUTAGEN., AIDS & NeurIPS'2020                                                                                           & {  \url{https://shorturl.at/OP015} }                                                                                                                                                                                                                                                                                         \\ \midrule

\begin{tabular}[c]{@{}l@{}}TAPool\tnote{[17]}\end{tabular}     & Graph Classification                                                                                & D\&D, PROTEINS, MUTAG, PTC & TPAMI'2021                                                                                           & --                                                                                                                                                                                                                                                                                              \\ \midrule

\begin{tabular}[c]{@{}l@{}}IPool\tnote{[18]}\end{tabular}        & Graph Classification                                                                                & D\&D, PROTEINS, ENZYMES, NCI1, NCI109, MNIST, CIFAR10 & TNNLS'2021                                                                                           & -- \\ \midrule

\multirow{4}{*}{MVPool\tnote{[19]}}  & Graph Classification & D\&D, PROTEINS,ENZYMES, NCI1, NCI109, MUTAGEN. IMDB-B, RDT-M12K &   \multirow{4}{*}{TKDE'2021}      &   \multirow{4}{*}{ \url{https://shorturl.at/fgqX1}}                   \\
                              & Node Classification    & Cora, Citeseer, Pubmed, Coauthor-CS, Coauthor-Phy & &  \\
                             & Node Clustering    & Cora, Citeseer, Pubmed & &  \\
                              & Graph Clustering   & PROTEINS, NCI109, MUTAGEN. & &\\ \midrule                                                                                                                                                                                                                                                                                 

\multirow{3}{*}{AdamGNN\tnote{[20]}}   & Graph Classification &  D\&D, PROTEINS, MUTAG NCI1, NCI109, MUTAGEN.  &   \multirow{3}{*}{TKDE'2022}      &   \multirow{3}{*}{ \url{https://shorturl.at/tNQWZ}}                   \\
                             & Node Classification    & Cora, Citeseer, Pubmed, DBLP, ACM, Emails, Wiki, Ogbn-arxiv & &  \\
                             & Link Prediction    & Cora, Citeseer, Pubmed, DBLP, ACM, Emails, Wiki & &        
\\ \bottomrule
\end{tabular}%
}
%\begin{tablenotes}   \tiny   %\scriptsize
%
% 
%  \item[1]   
%  “--” indicates that no open-source code is found.
%  \item[2] Models with \tnote{$\spadesuit$} have another implementation by Pytorch available in Pytorch Geometric:~\url{https://pytorch-geometric.readthedocs.io/}.
%  \item[3] Models with \tnote{$\clubsuit$} have another implementation by TensorFlow available in Spektral:~\url{https://graphneural.network/}.
%  \item[4] More up-to-date resource can be found at our repository~\url{https://github.com/LiuChuang0059/graph-pooling-papers}.
%  \item[5] ~\cite{sortpool}
%  \end{tablenotes}
  \end{threeparttable}

  \begin{minipage}{\linewidth}\scriptsize
$^{[1]}$\cite{sortpool}; $^{[2]}$\cite{second-order}; $^{[3]}$\cite{gmt}; $^{[4]}$\cite{dke-pool}; $^{[5]}$\cite{diffpool}; $^{[6]}$\cite{mincut}; $^{[7]}$\cite{mem-pool}; $^{[8]}$\cite{structpool}; $^{[9]}$\cite{hoscpool};  $^{[10]}$\cite{sep};   $^{[11]}$\cite{graph-u-net}; $^{[12]}$\cite{sagpool};  $^{[13]}$\cite{asap};  $^{[14]}$\cite{vip-pool}; $^{[15]}$\cite{gsapool}; $^{[16]}$\cite{path-pooling};  $^{[17]}$\cite{TAPool};  $^{[18]}$\cite{ipool}; $^{[19]}$\cite{MVpool}; $^{[20]}$\cite{adamgnn}.

\textbf{\textcircled{\raisebox{-0.9pt}{1}}} Models with $\spadesuit$ have another implementation by Pytorch available in Pytorch Geometric:~\url{https://pytorch-geometric.readthedocs.io/}. 

\textbf{\textcircled{\raisebox{-0.9pt}{2}}} Models with $\clubsuit$ have another implementation by TensorFlow available in Spektral:~\url{https://graphneural.network/}.
\end{minipage}
\caption{A list of representative graph pooling methods.}
\label{tab:models}
\end{table*}

\section{Libraries for Graph Pooling}
\label{sec:library}

\paragraph{Benchmark Datasets.} Table~\ref{tab:dataset} provides the statistics of commonly used datasets for evaluating graph pooling methods, which mainly come from two widely used repositories: TU dataset~\cite{tu-dataset}, which contains over 130 datasets varying in content domains and dataset sizes, and  Open Graph Benchmark (OGB) dataset~\cite{ogb-dataset}, which contains many large-scale benchmark datasets.  The above datasets can be classified into four categories:  \textbf{1) Social Networks.} The social networks consider entities as nodes, and their social interactions as edges. \textbf{2) Protein Networks.} The commonly used datasets include PROTEINS and D\&D, where nodes correspond to amino acids, and edges are constructed if the distance between two nodes is below 6 Angstroms. \textbf{3) Molecule Graphs.} The commonly used molecular datasets include NCI1 and MUTAG, in which nodes and edges refer to atoms and bonds, respectively. \textbf{4) Others.} Besides the above three types of datasets, there are also some less commonly used datasets, such as synthetic datasets introduced by Knyazev \textit{et al.}~\shortcite{understand-attention}, and image datasets~\cite{benchmarkgnns}, which are converted into graphs using super-pixels. The former is used to evaluate the generalization capability of pooling methods, and the latter is usually used to visualize the preserved information via graph pooling (usually used in node drop pooling methods), which helps analyze the interpretability of pooling methods.
%and fMRI datasets (HCP and ABIDE)~\cite{uniform-pooling}.

\paragraph{Model Architectures.} The graph pooling methods are generally evaluated by two levels of graph tasks: \textbf{1) Node-level tasks} include node classification and node clustering tasks, which are generally tested on a single graph in the form of transductive learning; \textbf{2) Graph-level tasks} include graph classification, graph regression, graph reconstruction, and graph generation tasks, which are usually tested on multiple graphs in the form of inductive learning. Figure~\ref{fig:model-arch} summarizes the model architectures commonly used for various tasks. The abbreviations used in the figure are MP for message passing, NC for node clustering, HP for hierarchical pooling, FP for flat pooling, and UP for unpooling. In addition to the above task-oriented evaluations, Daniele \textit{et al.}~\shortcite{understanding-pooling} provided another two evaluation criteria, including preserving the information content of the node attributes and preserving the topological structure, which help to comprehensively quantify the ability of graph pooling. 

\paragraph{Codes.} To facilitate the access to empirical analysis, we summarize the open-source codes of representative graph pooling studies in Table~\ref{tab:models}. Meanwhile, we list the applied tasks and corresponding benchmark datasets of each method. Due to space limitation,  a more complete summary (over 150 papers reviewed) is presented in our GitHub repository~\footnote{\url{https://github.com/LiuChuang0059/graph-pooling-papers}}. Moreover, we will update the repository in real-time as more methods and their implementations become available.

\section{Applications}
\label{sec:application}

We briefly review the recent studies that incorporate the idea of graph pooling on a wide range of applications, which can be divided into two classes according to the type of datasets:  \textbf{1) Structural datasets}, where the data have explicit relational structures, such as molecular property prediction~\cite{haar-pooling,mem-pool}, molecular generation~\cite{gmt}, protein-ligand binding affinity prediction~\cite{protein}, 3D protein structure analysis~\cite{3D-protein}, drug discovery~\cite{drug-discovery}, recommendation~\cite{user-as-graph,sequen-recommen,recommendation-gcn}, community detection~\cite{Community-Detection-survey-2}, and relation extraction of knowledge graph~\cite{KGPool}. \textbf{2) Non-structural datasets}, where the relational structures are implicit, and the graphs are built according to domain knowledge, such as cancer diagnosis~\cite{cancer,lung-cancer}, brain data analysis~\cite{pooling-regular,BrainGNN,brain-dynamic,asd,eeg,brain-analyze}, anti-spoofing and speech deepfaked detection~\cite{anti-spoofing}, natural language processing (NLP)~\cite{gpool-text,higil}, computer vision (CV)~\cite{ecc,splinecnn,object-recognition}, 3D point clouds~\cite{point-cloud,hapgn,papooling}, and multimodal sentiment analysis~\cite{multimodal}.
%and can be naturally performed in the graph structure,

The aforementioned studies have effectively decreased sample size~\cite{KGPool} and incorporated hierarchical information~\cite{hapgn} by utilizing either pre-existing or newly developed graph pooling techniques. This has resulted in notable enhancements across diverse applications, confirming the efficacy of graph pooling methods as demonstrated through experimental results in these works~\cite{user-as-graph,asd,L-GRIN,brain-analyze}. Thus, the applicability of graph pooling has been established.

\section{Challenges and Opportunities}
\label{sec:challenge}

%\uppercase\expandafter{\romannumeral3}

%\subsection{Different Tasks}

% \subsubsection{\uppercase\expandafter{\romannumeral1}: Different Tasks and Multi-tasks}

\subsection{Different Tasks and Multi-tasks}

\paragraph{Challenge.}  Table~\ref{tab:models} demonstrates that most existing graph pooling methods focus on graph-level tasks, with few addressing node-level tasks like node classification. Graph pooling methods show promise in reducing the number of parameters, which can decrease time complexity and increase model resistance to over-fitting. Additionally, these methods have large receptive fields that capture high-order information. However, designing efficient and effective  pooling/unpooling operators and integrating them into graph convolution networks for better performance in node-level tasks (\textit{e.g.}, node classification) are still  key challenges. 
% In addition, the relations between node-level tasks and graph-level tasks are still under exploration.

\paragraph{Opportunity.} Some methods attempt to handle the node classification task with an encoder-decoder learning structure as shown in Figure~\ref{fig:model-arch} (e), where the unpooling operation is an important component. Several node drop pooling methods~\cite{vip-pool,MVpool} utilize the unpooling operation proposed in the Graph U-net~\cite{graph-u-net}. Another important component is the convolution of decoder, termed deconvolution. Besides directly adopting the convolution of encoder as deconvolution, we can design a more reasonable deconvolution, inspired by DGN~\cite{deconvolutional}, which reconstructs graph signals from smoothed node representations from the perspective of spectral domain. Therefore, future research directions include: 1) handling node-level tasks with an encoder-decoder structure by designing efficient unpooling and deconvolutional operators;  2) developing a unified graph pooling approach, through which node-level and graph-level tasks can be simultaneously handled~\cite{multi-task,shadow}, which is beneficial since different tasks may benefit from each other.

\subsection{Different Types of Graphs}
\paragraph{Challenge.}  The graph pooling methods discussed in Section~\ref{sec:methods} are primarily intended for plain graphs. However, there are also many other types of graphs in real-world datasets. Obviously, it is not optimal to directly apply the above existing graph pooling methods to these different types of graphs because they possess distinct characteristics. Therefore, there is a significant gap in the literature regarding the development of specific graph pooling operators tailored to handle diverse types of graphs.

\paragraph{Opportunity.} Recently, several graph pooling works have attempted to handle the rarely studied but commonly used graphs in real applications, such as heterogenous graphs~\cite{user-as-graph}, spatio-temporal graphs~\cite{spatial-temporal}, and hypergraphs~\cite{hyper-pool}. However, specific graph pooling operators for handling other types of graphs are still lacking.  Therefore, we suggest two promising research opportunities: 1) extending the existing graph pooling methods or designing new pooling methods to deal with specific graphs in consideration of their properties, such as dynamic graphs, where graph structures and graph nodes dynamically change over time, directed graphs, where the edges have a direction, and signed networks, where signed edges represent positive or negative relationships between nodes; 2) devising a general pooling method, which can efficiently handle different types of graphs in a universal manner.

% \subsubsection{\uppercase\expandafter{\romannumeral3}: Interpretability}

\subsection{Interpretability}
\paragraph{Challenge.}  Although the existing graph pooling methods have achieved excellent results on various graph-based tasks, most of them show little interpretability~\cite{understand-attention}. This is particularly problematic for drug or disease-related problems where understanding the reasoning behind these methods is crucial. Although there has been some progress in interpreting node representation of GNNs~\cite{explaina-survey}, the interpretability of pooling methods remains largely unexplored.

% However, the interpretability is desirable and critical for pooling methods on graphs, especially in terms of drug or disease related problems, since it sheds light on the decision made by these methods. However, despite some progress made in the interpretation of node representation via GNNs~\cite{explaina-survey}, the interpretability of pooling methods remains underexplored.

\paragraph{Opportunity.} To exactly explain what has been learned from graph pooling operations, some studies~\cite{diffpool,mem-pool,commpool} have attempted to present the visualization of hierarchical clusters or hierarchical community structures captured by pooling operations, but they do not provide quantitative analyses to assess the quality of clustering. Moreover, the interpretability of pooling operations is still not well explored. Therefore, there is a need for further research in this area. A promising direction would be to extend existing studies on interpreting GNNs~\cite{explaina-survey} to include graph pooling.

% Thus, generalizing the studies on the interpretation of GNNs~\cite{explaina-survey} into graph pooling may be a promising future research direction.

% \subsubsection{\uppercase\expandafter{\romannumeral4}: Robustness}

\subsection{Robustness}
\paragraph{Challenge.}  Since many applications of graph pooling methods are risk-sensitive, \textit{e.g.}, drug design and disease diagnosis, the robustness of methods is essential and indispensable for actual usages. However, according to the analysis in recent studies~\cite{adversarialpool,HIBpool,liu2022on}, most graph pooling methods fail to distinguish the noise information from the input graph, thus dramatically degrading their performance when the input graph is perturbed in terms of features or topology. 

\paragraph{Opportunity.} Although some initial studies have been conducted on the robustness of graph machine learning, few studies have explored the robustness of graph pooling methods. So far, only an adversarial attack framework has been proposed by Tang \textit{ et al.}~\shortcite{adversarialpool} to evaluate the robustness of graph pooling methods. Therefore, there is much work to be done to develop a robust graph pooling method for practical applications, such as: 1) building up a comprehensive  adversarial attack framework encompasses various types of attacks; 2) designing adversarial defense graph pooling methods. One potential solution is to extend the techniques used for enhancing the robustness of graph machine learning into improving the robustness of graph pooling methods.

% NCPool~\cite{node-clustering} is devised to capture local graph information, which can effectively enhance the robustness of graph clustering for the spurious edges.

% \subsubsection{\uppercase\expandafter{\romannumeral5}: Large-scale Data}

\subsection{Large-scale Data}
\paragraph{Challenge.}  Most graph pooling methods are tested on small benchmark datasets, which may be insufficient for comparisons among different graph pooling models. For example, applying graph pooling methods to small node classification datasets, \textit{i.e.}, Cora, is insufficient to assess their effectiveness in reducing the time and space complexity. Only a few works have attempted to deal with relatively larger graph datasets~\cite{adamgnn}. Additionally, the efficiency of pooling methods is crucial. The high cost of time or space limits their applicability to large-scale datasets. Node clustering pooling methods, for example, suffer from high storage complexities due to the computation of dense assignment matrices~\cite{understanding-pooling}.

\paragraph{Opportunity.} Hence, we suggest the following research objectives: 1) performing further verifications on large-scale graph datasets, \textit{e.g.}, on the Open Graph Benchmark~\cite{ogb-dataset}, which is a recently proposed large-scale graph machine learning benchmark;  2) designing more efficient graph pooling methods and making them more practical with constrained resource in real-world scenarios.

% \subsubsection{\uppercase\expandafter{\romannumeral6}: Expressive Power}

\subsection{Expressive Power}
\paragraph{Challenge.}  Most existing graph pooling methods are designed by intuition, and their performance gains are evaluated by empirical experiments. The lack of the means by which we can characterize the expressive power of graph pooling operators hinders the creation of more powerful graph pooling models~\cite{gin}.

\paragraph{Opportunity.} Only a limited number of studies~\cite{relation-pooling,LRP,gmt} have examined the expressive ability of their models in terms of the 1-Weisfeiler-Lehman (WL) test. Recently, Bianchi \textit{et al.}~\shortcite{expressive-pooling} conducted an comprehensive analysis on the expressive power of graph pooling techniques.  Therefore, based on their theoretical findings, it is promising and significant to explore more powerful graph pooling methods as future research directions. 

% \subsubsection{\uppercase\expandafter{\romannumeral7}: Generalize to Out-of-Distribution Data.}

\subsection{Generalize to Out-of-Distribution Data.}
\paragraph{Challenge.}  Graph Neural Networks are proposed without considering the agnostic distribution shifts between training graphs and testing graphs, causing the degeneration of the generalization ability in out-of-distribution (OOD) settings. Boris \textit{et al.}~\cite{understand-attention} have attempted to improve the generalization ability of GNNs with the help of graph pooling models. Moreover, Xu \textit{et al.}~\cite{extrapolate} emphasized the significance of selecting appropriate pooling functions for enabling GNNs to generalize over graph data beyond the distribution of training data.

\paragraph{Opportunity.} Recently, to improve the OOD generalization ability of GNNs has become an appealing and non-trivial task. According to the above introduction, utilizing graph pooling techniques can be an effective approach to improve the generalization capability of GNNs on OOD graph data.  

\section*{Acknowledgments}

This work was supported in part by the Natural Science Foundation of China (Nos. 61976162, 82174230, 62002090), Artificial Intelligence Innovation Project of Wuhan Science and Technology Bureau (No. 2022010702040070), and Science and Technology Major Project of Hubei Province (Next Generation AI Technologies) (No. 2019AEA170). Dr Wu is partially supported by ARC Projects LP210301259 and DP230100899. Prof Dacheng Tao is partially supported by Australian Research Council Project FL-170100117.

% \clearpage
% \small
%\footnotesize
%\scriptsize
%\tiny

\bibliographystyle{ijcai22}
\bibliography{ijcai22}

%
%%% The file named.bst is a bibliography style file for BibTeX 0.99c
%\bibliographystyle{named}
%\bibliography{ijcai23}

\end{document}